\documentclass[runningheads]{llncs}
\usepackage{graphicx}
\begin{document}
\title{Workspace Analysis for Laparoscopic Rectal Surgery :
A Preliminary Study}
\titlerunning{Workspace Analysis for Laparoscopic Rectal Surgery : A Preliminary Study}
%
\author{Alexandra Thomieres\inst{1} \and
Dhruva Khanzode\inst{2,3,4}\orcidID{0000-0003-4207-1221} \and
Emilie Duchalais\inst{1}\orcidID{0000-0003-3654-2963}\and
Ranjan Jha\inst{2,3}\orcidID{0000-0002-4741-0475}\and
Damien Chablat\inst{4,5}\orcidID{0000-0001-7847-6162}}
\authorrunning{A. Thomieres et al.}
%
\institute{Centre Hospitalier Universitaire, Nantes 44000, France \email{\{alexandra.thomieres,emilie.dassonneville\}@chu-nantes.fr} \and
CSIR-Central Scientific Instruments Organisation, Chandigarh 160030, India
\email{ranjan.jha@csio.res.in} \and
Academy of Scientific and Innovative Research (AcSIR), Ghaziabad 201002, India \email{dhruva.csio20a@acsir.res.in}\and École Centrale Nantes, Nantes Université, CNRS, LS2N, UMR 6004,
F-44000  Nantes, France \\
\email{damien.chablat@cnrs.fr} \and 
Research Center for Industrial Robots Simulation and Testing, Technical University of Cluj-Napoca, 400114 Cluj-Napoca, Romania \\
}
\maketitle              
\begin{abstract} 
The integration of medical imaging, computational analysis, and robotic technology has brought about a significant transformation in minimally invasive surgical procedures, particularly in the realm of laparoscopic rectal surgery (LRS). This specialized surgical technique, aimed at addressing rectal cancer, requires an in-depth comprehension of the spatial dynamics within the narrow space of the pelvis. Leveraging Magnetic Resonance Imaging (MRI) scans as a foundational dataset, this study incorporates them into Computer-Aided Design (CAD) software to generate precise three-dimensional (3D) reconstructions of the patient's anatomy. At the core of this research is the analysis of the surgical workspace, a critical aspect in the optimization of robotic interventions. Sophisticated computational algorithms process MRI data within the CAD environment, meticulously calculating the dimensions and contours of the pelvic internal regions. The outcome is a nuanced understanding of both viable and restricted zones during LRS, taking into account factors such as curvature, diameter variations, and potential obstacles. This paper delves deeply into the complexities of workspace analysis for robotic LRS, illustrating the seamless collaboration between medical imaging, CAD software, and surgical robotics. Through this interdisciplinary approach, the study aims to surpass traditional surgical methodologies, offering novel insights for a paradigm shift in optimizing robotic interventions within the complex environment of the pelvis.

\keywords{Laparoscopic rectal surgery \and Magnetic Resonance
Imaging  \and Workspace analysis}
\end{abstract}
%

\section{Introduction}

In the field of minimally invasive surgical procedures, the fusion of medical imaging, computational analysis, and robotic technology has significantly advanced the precision and efficacy of the procedures. Despite the implementation of robotic assistance, laparoscopic rectal surgery (LRS) remains one of the most difficult procedure due to the narrowness of the workspace into the pelvis. Understanding the spatial dynamics and complexity of the human pelvic workspace is important for optimizing the performance of robotic systems used in these procedures. This requires a comprehensive workspace analysis, where Magnetic Resonance Imaging (MRI) scans serve as the foundational data source, providing detailed anatomical information.

The use of pelvic MRI before LRS has progressively become the gold standard of analysis, surpassing other techniques like endorectal ultrasound and CT-scanning.  \cite{brown2004effectiveness,jhaveri2009role}. Over the past two decades, the consistent improvement in MRI quality has positioned it as one of the most effective examination procedures for the anatomical analysis of pelvic tissues. 
The integration of MRI slices into Computer-Aided Design (CAD) software could stand as a pivotal step in understanding the complex geometry of the pelvis. This integration allows for the creation of three-dimensional (3D) reconstructions that faithfully replicate the patient's anatomy. These reconstructions, derived from high-resolution MRI scans, serve as the virtual models upon which the subsequent workspace analysis could be carried out for robotic surgical equipment and tools.

The workspace analysis, a critical feature of this study, involves differentiating the feasible and restricted zones within the pelvis that can be crossed by robotic instruments during a LRS. By employing advanced computational algorithms within the CAD environment, the software processes the comprehensive MRI data to precisely calculate the dimensions and contours of the pelvis area. This computational procedure facilitates a better understanding of the workspace, considering factors such as dimensions, curvature, diameter variations, and potential obstacles.

This paper describes the methodology of the fusion of medical imaging, CAD modeling, and robotic apparatus utilising raw MRI data from the patients and using it to generate a virtual 3D CAD model. The aim is to facilitate the analysis and optimisation of available workspace for a robotic surgical tool during minimally invasive laparoscopic surgery \cite{khanzode2023surgical,pisla2009kinematical}.
\section{Medical Imaging}
Medical imaging allows us for a clear visualization of the muscles of the pelvic floor, blood vessels, and a distinct identification of the soft tissues surrounding the rectum \cite{Hong2020}. Numerous studies have investigated whether 2D pelvimetry could predict the difficulty of total mesorectal excision in patients operated for rectal resection \cite{Zhang2023}. The narrowness of the pelvis can be estimated by several measurements obtained from MRI imaging, including obstetric conjugate, pelvic depth, sacral curvature angle, sacral depth, transverse diameter, interspinous distance, and intertuberous distance. These measurements are associated with intraoperative difficulty, in terms of operative duration or intraoperative blood loss \cite{Akiyoshi2009,Atasoy2018,Kim2011,Killeen2010,Teng2022}.  

Three-dimensional reconstruction could improve pelvic evaluation and facilitate pelvimetry and is frequently performed in gynecology and urology, but it remains uncommon in colorectal surgery \cite{Lenhard2009,Doumouchtsis2017}. Virzi et al. explored the mapping of pelvic tumors in children using 3D Slicer software  \cite{Virzi2020}. It could provide improved visibility of the pelvic space by delineating the bony structure, soft tissues, and the genitourinary system. Such visualization could enhance the understanding of challenges encountered during surgery and better estimate the area available to maneuver surgical instruments. 

\section{Methodology}
This retrospective study included patients with rectal cancer (within 15 cm of the anal verge) who had undergone preoperative MRI with the T2 sequence available on our software and surgical resection by laparoscopy between 2020 and 2023. Patients were informed that their data could be used for research.

\begin{figure}[!ht]
    \centering
    \includegraphics[width=\textwidth]{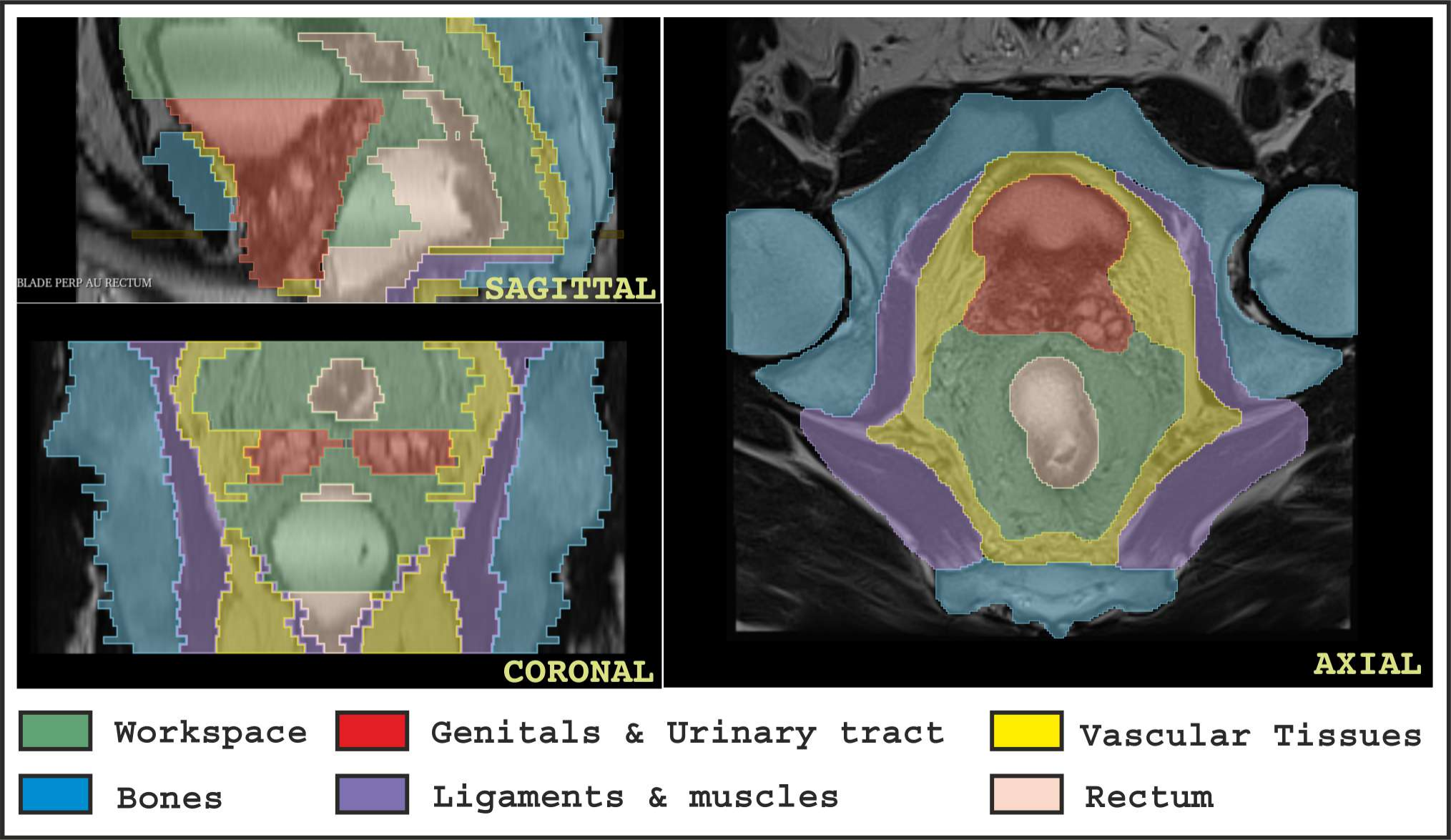}
    \caption{Marked MRI with differentiation of tissues in Axial, coronal, and Sagittal plane}
    \label{fig:MRI}
\end{figure}

{\it Radiological study: } 
For pelvic and rectal imaging, T2-weighted MRI images were acquired in the Axial, Coronal, and Sagittal planes. The images were exported using the CHU Nantes radiology software, CARESTREAM Vue PACS (version 12.1.6.1005, CARESTREAM Health, New York, USA). 
For pelvic volumetry, DICOM data from selected patients were analyzed using open-source software, 3D Slicer image computing platform (version 5.4.0). 
In the Axial section, the region studied is delimited cranially by the promontory and caudally by the top of the anal canal. The distal limit of stapling was located 1 cm above the superior pole of the internal sphincter.

The pelvic cavity was divided into six zones, as depicted in Figure~\ref{fig:MRI}:
\begin{itemize}
\item The working area (green), is the most accessible, the stapler can run smoothly by compressing the fatty mesorectal tissue.
\item Bones (blue), are hard and incompressible in nature.
\item Genito-urinary system (red): The bladder was considered empty due to the presence of a urinary catheter during surgery. In men, prostate and seminal vesicles were included in this zone. In women, since the uterus can be fully mobilized anteriorely during surgery, we opted to neglect it and instead used the cervix to delineate the upper part of the female genitourinary tract.
\item Vasculo-fatty tissues (yellow) encompass the ureters, iliac vessels, pre-sacral fascia, and certain nerves. This area exhibits low compressibility.
\item Muscles and ligaments (purple) are hardly compressible.
\item The rectum is represented in the color beige. In some MRI, rectal gel was used to facilitate the intraluminal visualization of the tumor. To mitigate any bias associated with rectal filling, we intentionally reduced the size of the rectum to mimick an empty rectum (no intraluminal space) as during surgery.
\end{itemize}

\begin{figure}[!ht]
    \centering
    \includegraphics[scale=0.145]{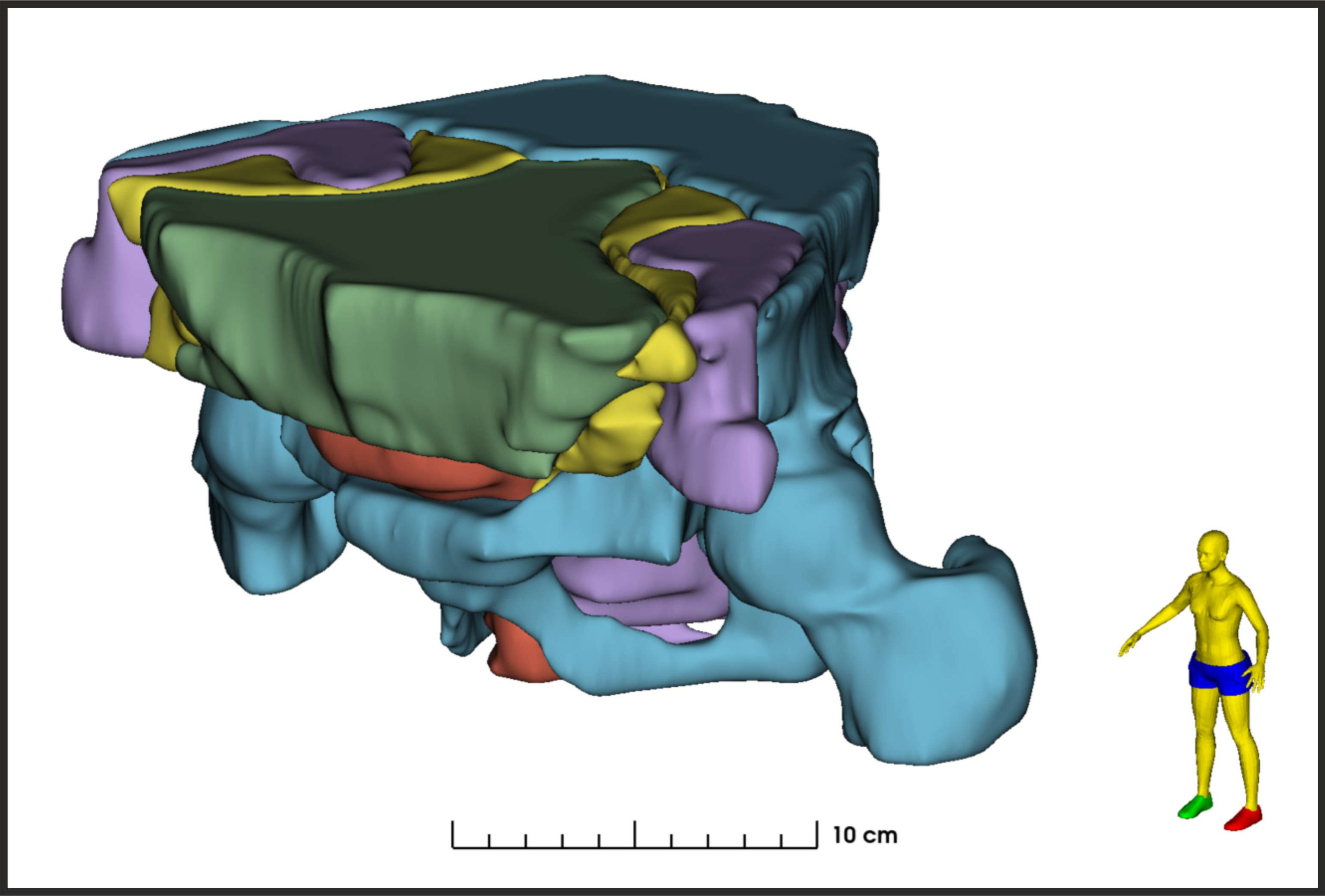}
    \caption{3D representation of pelvic cavity using MRI data showing all the six zones.}
    \label{fig:3d MRI}
\end{figure}
\begin{figure}
    \centering
    \includegraphics[scale=0.145]{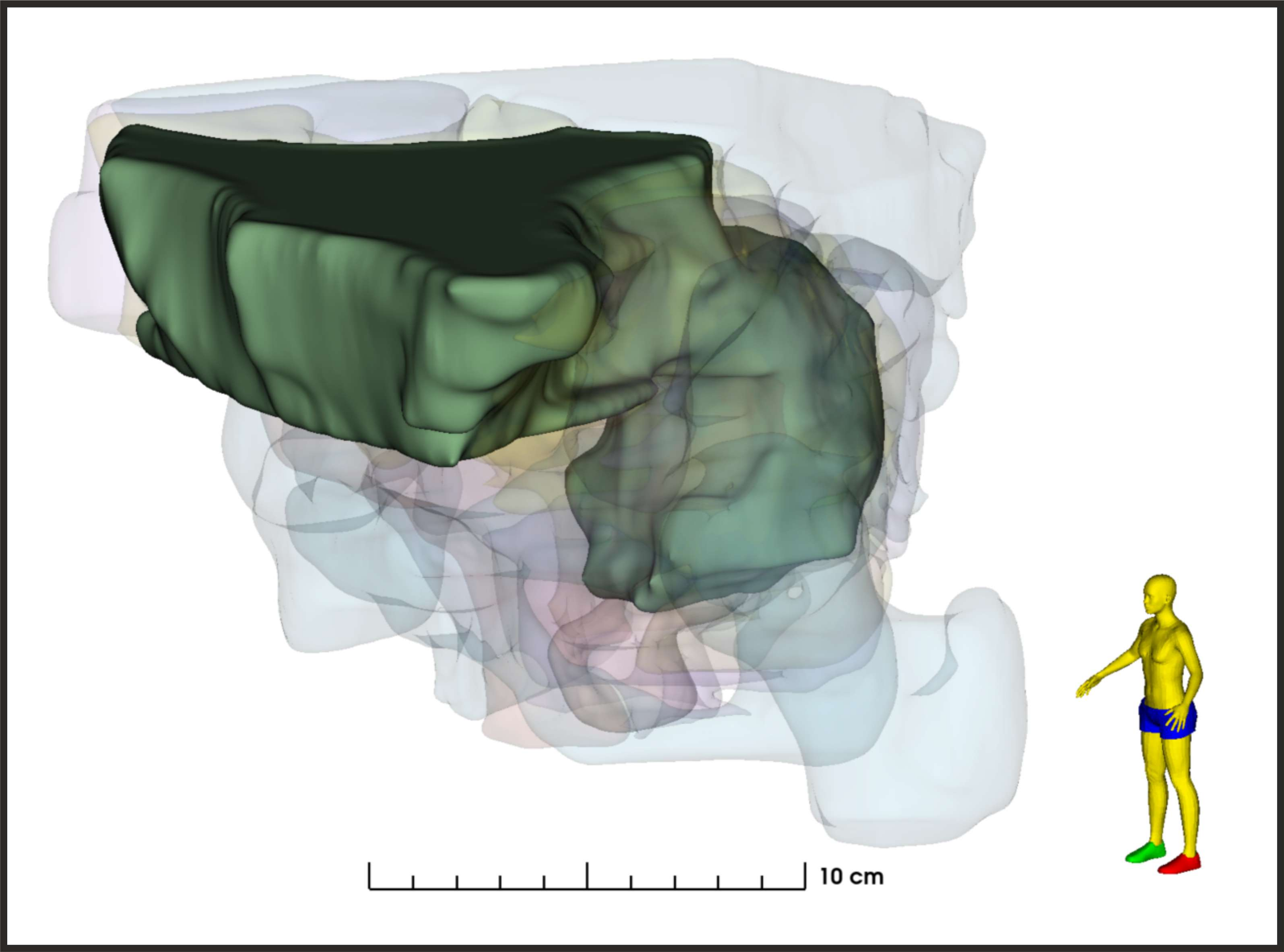}
    \caption{3D representation of pelvic MRI data highlighting solely the working area}
    \label{fig:MRI workspace}
\end{figure}
The tumour was not separately delimited but rather included within the rectum area. Contours of the areas were manually traced on all available axial images only for each patient. After axial delimitation, the contours were reviewed and adjusted as needed in sagittal and coronal planes. Section thickness varied among patients, ranging between 3 mm and 4.20 mm based on MRI specifications. The overall data was analyzed based on the patient's gender and BMI of the particular patient.

Subsequently, a 3D reconstruction of the pelvic cavity based on the marked MRI data was generated, as represented in Figure~\ref{fig:3d MRI}, and the workspace area (green) was isolated within the cluster of various marked zones, as depicted in Figure~\ref{fig:MRI workspace}. The data was then exported in the CAD-compatible \textit{.stl} format for analysis by the experienced design engineers.

\section{Workspace analysis}
In this section, the \textit{.stl} files generated in the previous step are put to work. In this step, we use Computer-Aided Design (CAD) software, SOLIDWORKS (version 2023, Dassault systems, Paris, France) to carry out design analysis and workspace evaluation.

\begin{figure}
    \centering
    \includegraphics[width=\textwidth]{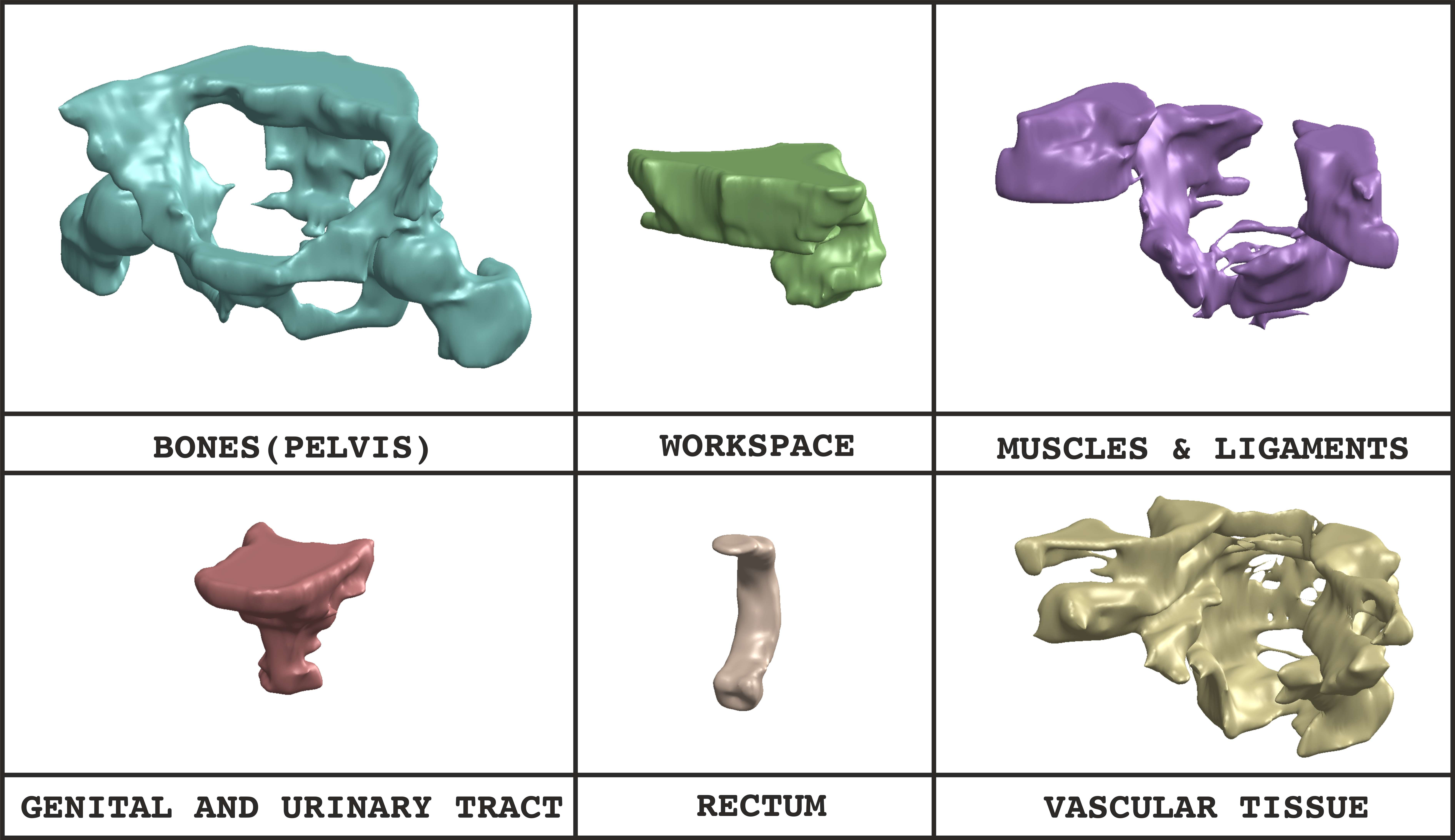}
    \caption{3D CAD models depicting the six distinct zones within the surgical site}
    \label{fig:CAD parts}
\end{figure}

In the virtual design domain, the \textit{.stl} files are imported into SOLIDWORKS and are converted into a modifiable \textit{.sldprt} file format. Once the file is in \textit{.sldprt} format, we can perform various design operations, feature alterations, and dimensional analysis in a much effortless manner. In the previous step, when \textit{.stl} files were generated for each and every tissue type, like Bones, Ligament, Vascular Tissue etc. that each file is to be imported and inspected carefully, for some isolated parts that needed to be removed before analysis. The 3D digital models of each and every tissue type in SOLIDWORKS are depicted in Figure~\ref{fig:CAD parts}. Once all the parts are inspected and converted successfully, all parts are to be combined to generate a mechanical assembly of the system. After the assembly is successfully carried out, we need to compare it with the 3D representation generated by the 3D Slicer software in the previous step. 

\begin{figure}[!ht]
    \centering
    \includegraphics[width=\textwidth]{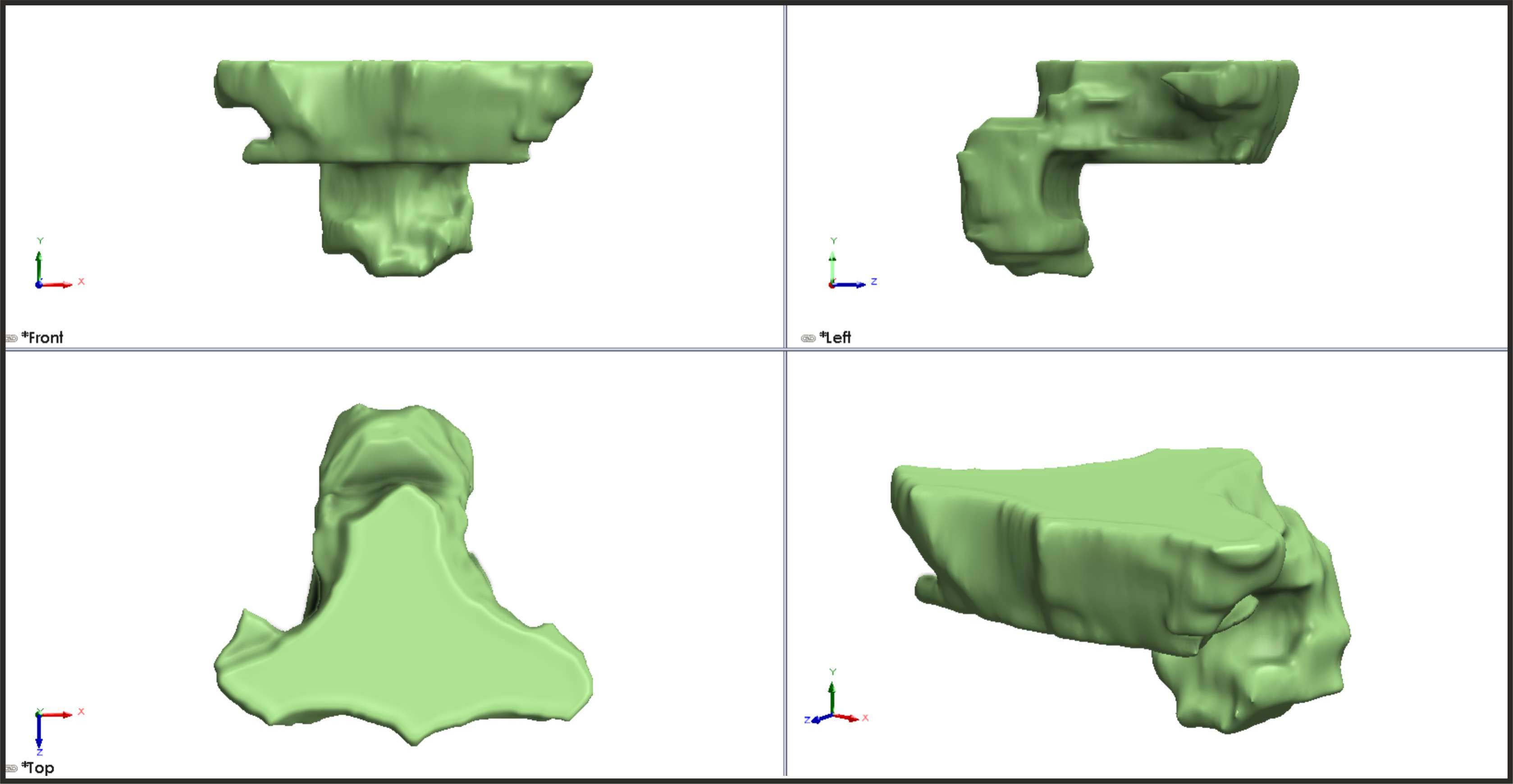}
    \caption{Front, Left, Top, and Isometric views of 3D workspace from CAD software.}
    \label{fig:3d workspace cad}
\end{figure}

After this step, we mainly focus on the ``workspace'' or the "working area" 3D CAD model as depicted in Figure~\ref{fig:3d workspace cad}. The internal dimensional tools of SOLIDWORKS software can be used to calculate the dimensional data such as centre of mass, overall surface area, and total volume of the model body. This dimensional data, specifically the volumetric data can be utilized to determine the workspace for the robotic tool during minimally invasive surgical procedures. 
The dimensional and structural data obtained provided us with valuable insights into anatomical constraints and variations in the workspace due to differences in patient types. In this study, we included both male and female patients with diverse body shape. Our observations highlighted that, from a workspace perspective, the total available workspace is greater in female patients than in male patients, suggesting anatomical constraints that differ between genders. Furthermore, we concluded that the actual site for maneuvering the robotic surgical tool is constrained to the lower half of the total available free workspace, and the shape of the functional workspace is similar to a concentric cylindrical structure.
\section{Conclusions and future work}
In summary, this article presents a methodology that involves utilizing raw MRI data to construct a functional 3D CAD model. This process includes precise tissue identification, marking, and data refinement to develop a CAD model that can be analyzed both dimensionally and volumetrically. Additionally, our preliminary findings indicate a greater total workspace in female patients compared to male patients, although other physiological parameters also probably contribute to workspace availability. Looking ahead, our future plans will involve studying of the anatomical properties of 40 patients representative of the epidemiological data for both female and male groups in our hospital.
Then, a more in-depth analysis of the CAD model will be conducted to precisely determine the available workspace for robotic surgical tools near the surgical site and allow the surgeon to anticipate the surgical strategy. This analysis will contribute to the development of a robotic surgical tool specifically designed to facilitate minimally invasive LRS.
\section*{Acknowledgments}
This research was supported by CSIR-Central Scientific Instruments Organization, Chandigarh, India, Centre Hospitalier Universitaire de Nantes, and Centrale Nantes, France. This research is funded by the Indian Council of Medical Research, New Delhi, India under the ``Senior Research Fellowship'' (File no.5/3/8/46/ITR-F/2022) and Centrale Nantes, Nantes, France under Contrat doctoral privé (décret 2021-1233) awarded to Mr. Dhruva Rajesh Khanzode \& by AMI CHU /Centrale Nantes financing, awarded to Ms. Alexandra Thomieres.  This research was supported by the project New Smart and Adaptive Robotics Solutions for Personalized Minimally Invasive Surgery in Cancer Treatment - ATHENA, funded by the European Union – NextGenerationEU and Romanian Government, under National Recovery and Resilience Plan for Romania, contract no. 760072 /23.05.2023, code CF 16/15.11.2022, through the Romanian Ministry of Research, Innovation and Digitalization, within Component 9, investment I8.
%
%
%
\bibliographystyle{splncs04}
\bibliography{reference.bib}
\end{document}